\def\year{2016}
\documentclass[letterpaper]{article}
\usepackage{aaai16}
\usepackage{times}
\usepackage{helvet}
\usepackage{courier}
\usepackage{array}
\usepackage{multirow}
\usepackage{amsmath}
\usepackage{amssymb}
\usepackage{graphicx}

\usepackage{floatrow}
\DeclareFloatFont{tiny}{\tiny}
\newcommand{\noun}[1]{\textsc{#1}}
\providecommand{\tabularnewline}{\\}

\pubnote{\em The Workshops of the Thirtieth AAAI Conference on Artificial Intelligence}
\begin{document}

\title{Deep Activity Recognition Models with Triaxial Accelerometers}
\author{{Mohammad Abu Alsheikh \and Ahmed Selim \and Dusit Niyato} \AND {\bf Linda Doyle \and Shaowei Lin \and Hwee-Pink Tan}
}
\maketitle
\begin{abstract}
\begin{quote}
Despite the widespread installation of accelerometers in almost all mobile phones and wearable devices, activity recognition using accelerometers is still immature due to the poor recognition accuracy of existing recognition methods and the scarcity of labeled training data. We consider the problem of human activity recognition using triaxial accelerometers and deep learning paradigms. This paper shows that deep activity recognition models (a)~provide better recognition accuracy of human activities, (b)~avoid the expensive design of handcrafted features in existing systems, and (c)~utilize the massive unlabeled acceleration samples for unsupervised feature extraction. Moreover, a hybrid approach of deep learning and hidden Markov models (DL-HMM) is presented for sequential activity recognition. This hybrid approach integrates the hierarchical representations of deep activity recognition models with the stochastic modeling of temporal sequences in the hidden Markov models. We show substantial recognition improvement on real world datasets over state-of-the-art methods of human activity recognition using triaxial accelerometers.
\end{quote}
\end{abstract}

\section{Introduction}


Existing sensor-based activity recognition systems~\cite{chen2012sensor} use shallow and conventional supervised machine learning algorithms such as multilayer perceptrons (MLPs), support vector machines, and decision trees. This reveals a gap between the recent developments of deep learning algorithms and existing sensor-based activity recognition systems. When deep learning is applied for sensor-based activity recognition, it results in many advantages in terms of system performance and flexibility. Firstly, deep learning provides an effective tool for extracting high-level feature hierarchies from high-dimensional data which is useful in classification and regression tasks~\cite{salakhutdinov2015learning}. These automatically generated features eliminate the need for handcrafted features of existing activity recognition systems. Secondly, deep generative models, such as deep belief networks~\cite{hinton2006fast}, can utilize unlabeled activity samples for model fitting in an unsupervised pre-training phase which is exceptionally important due to the scarcity of labeled activity datasets. In the contrary, unlabeled activity datasets are abundant and cheap to collect. Thirdly, deep generative models are more robust against the overfitting problem as compared to discriminative models (e.g., MLP)~\cite{mohamed2012acoustic}.

In this paper, we present a systematic approach towards detecting human activities using deep learning and triaxial accelerometers. This paper is also motivated by the success of deep learning in acoustic modeling~\cite{mohamed2012acoustic,dahl2012context}, as we believe that speech and acceleration data have similar patterns of temporal fluctuations. Our approach is grounded over the automated ability of deep activity recognition models in extracting intrinsic features from acceleration data. Our extensive experiments are based on three public and community-based datasets. In summary, our main results on deep activity recognition models can be summarized as follows:
\begin{itemize}
\item \textbf{Deep versus shallow models}. Our experimentation shows that using deep activity recognition models significantly enhances the recognition accuracy compared with conventional shallow models. Equally important, deep activity recognition models automatically learn meaningful features and eliminate the need for the hand-engineering of features, e.g., statistical features, in state-of-the-art methods.
\item \textbf{Semi-supervised learning}. The scarce availability of labeled activity data motivates the exploration of semi-supervised learning techniques for a better fitting of activity classifiers. Our experiments show the importance of the generative (unsupervised) training of deep activity recognition models in weight tuning and optimization.
\item \textbf{Spectrogram analysis}. Accelerometers generate multi-frequency, aperiodic, and fluctuating signals which complicate the activity recognition using time series data. We show that using spectrogram signals instead of the raw acceleration data exceptionally helps the deep activity recognition models to capture variations in the input data.
\item \textbf{Temporal Modeling}. This paper presents a hybrid approach of deep learning and hidden Markov model (DL-HMM) for better recognition accuracy of temporal sequence of activities, e.g., fitness movement and car maintenance checklist. This hybrid technique integrates the hierarchical representations of deep learning with stochastic modeling of temporal sequences in HMMs. Experiments show that a DL-HMM outperforms HMM-based methods for temporal activity recognition. Specifically, the learned representation of deep activity recognition models is shown to be effective in estimating the posterior probabilities of HMMs. Unlike Gaussian mixture models which provide an alternative method, deep neural networks do not impose restrict assumptions on the input data distribution~\cite{mohamed2012acoustic}.
\end{itemize}

\section{Related Work}\label{sec:related_work}

In this section, we will focus on classification and feature engineering methods for activity recognition using accelerometers. For a more comprehensive review of the field, we refer interested readers to recent survey papers~\cite{lara2013survey,chen2012sensor}.

\subsection{Limitations of Shallow Classifiers}\label{sub:limitations_shallow_classifiers}

Machine learning algorithms have been used for a wide range of activity recognition applications~\cite{parkka2006activity,khan2010triaxial,altun2010human,kwapisz2011activity}, allowing the mapping between feature sets and various human activities. The classification of accelerometer samples into static and dynamic activities using MLPs is presented in~\cite{khan2010triaxial}. Conventional neural networks, including MLPs, often stuck in local optima~\cite{rumelhart1988learning} which leads to poor performance of activity recognition systems. Moreover, training MLPs using backpropagation~\cite{rumelhart1988learning} only hinders the addition of many hidden layers due to the vanishing gradient problem. The authors in~\cite{parkka2006activity} used decision trees and MLPs to classify daily human activities. In~\cite{berchtold2010actiserv}, a fuzzy inference system is designed to detect human activities. \cite{kwapisz2011activity} compared the recognition accuracy of decision tree (C4.5), logistic regression, and MLPs, where MLPs are found to outperform the other methods.

In this paper, we show significant recognition accuracy improvement on real world datasets over state-of-the-art methods for human activity recognition using triaxial accelerometers. Additionally, even though some previous works have purportedly reported promising results of activity recognition accuracy, they still require a degree of handcrafted features as discussed below.

\subsection{Limitations of Handcrafted Features}\label{sub:limitations_handcrafted}

Handcrafted features are widely utilized in existing activity recognition systems for generating distinctive features that are fed to classifiers. The authors in~\cite{altun2010human,berchtold2010actiserv,kwapisz2011activity,xu2012robust,catal2015use} utilized statistical features, e.g., mean, variance, kurtosis and entropy, as distinctive representation features. On the negative side, statistical features are problem-specific, and they poorly generalize to other problem domains. In~\cite{zappi2007activity}, the signs of raw signal (positive, negative, or null) are used as distinctive features. Despite its simple design, these sign features are plain and cannot represent complex underlying activities which increase the number of required accelerometer nodes. The authors in~\cite{bachlin2010wearable} used the energy and frequency bands in detecting the freezing events of Parkinson\textquoteright s disease patients. Generally speaking, any handcrafted-based approach involves laborious human intervention for selecting the most effective features and decision thresholds from sensory data.

Quite the contrary, data-driven approaches, e.g., using deep learning, can learn discriminative features from historical data which is both systematic and automatic. Therefore, deep learning can play a key role in developing self-configurable framework for human activity recognition. The author in~\cite{plotz2011feature} discussed the utilization of a few feature learning methods, including deep learning, in activity recognition systems. Nonetheless, this prior work is elementary in its use of deep learning methods, and it does not provide any analysis of the deep network construction, e.g., setup of layers and neurons. Moreover, our probabilistic framework supports temporal sequence modeling of activities by producing the activity membership probabilities as the emission matrix of an HMM. This is a considerable advantage for temporally modeling human actions that consist of a sequence of ordered activities, e.g., fitness movement and car maintenance checklist.

\section{Problem Statement}\label{sec:system_model}

This section gives a formal description of the activity recognition problem using accelerometer sensors.

\subsection{Data Acquisition}


Consider an accelerometer sensor that is attached to a human body and takes samples (at time index $t$ ) of the form
\begin{equation}
\mathbf{r}_{t}=\mathbf{r}_{t}^{*}+\mathbf{w}_{t},\quad t=1,2,\ldots
\end{equation}
where $\mathbf{r}_{t}=\left[\begin{array}{ccc} r_{t}^{x} & r_{t}^{y} & r_{t}^{z}\end{array}\right]^{T}$ is a 3D accelerometer data point generated at time $t$ and composed of $r_{t}^{x}$, $r_{t}^{y}$, and $r_{t}^{z}$ which are the x-acceleration, y-acceleration, and z-acceleration components, respectively. The proper acceleration in each axis channel is a floating-point value that is bounded to some known constant $B>0$ such that $\left|r_{t}^{x}\right|\leq B$, $\left|r_{t}^{y}\right|\leq B$, and $\left|r_{t}^{z}\right|\leq B$. For example, an accelerometer with $B=2\text{g}$ units indicates that it can record proper acceleration up to twice the gravitational acceleration (recall that $1\text{g}\simeq9.8\tfrac{meter}{second^{2}}$). Clearly, an accelerometer that is placed on a flat surface record a vertical acceleration value of $\pm1\text{g}$ upward. $\mathbf{r}_{t}^{*}\in\mathbb{R}^{3}$ is a vector that contains 3-axial noiseless acceleration readings. $\mathbf{w}_{t}\in\mathbb{R}^{3}$ is a noise vector of independent, zero-mean Gaussian random variables with variance $\sigma_{w}^{2}$ such that $\mathbf{w}_{t}\backsim\mathcal{N}(0,\sigma_{w}^{2}\mathbb{I}_{3})$. Examples of added noise during signal acquisition include the effect of temperature drifts and electromagnetic fields on electrical accelerometers~\cite{fender2008two}.

Three channel frames $\mathbf{s}_{t}^{x}$, $\mathbf{s}_{t}^{y}$, and $\mathbf{s}_{t}^{z}\in\mathbb{R}^{\text{N}}$ are then formed to contain the x-acceleration, y-acceleration, and z-acceleration components, respectively. Particularly, these channel frames are created using a sliding window as follows:
\begin{eqnarray}
\mathbf{s}_{t}^{x} & = & [\begin{array}{ccc}
r_{t}^{x} & \cdots & r_{t+N-1}^{x}\end{array}]^{T},\label{eq:x_acc_time}\\
\mathbf{s}_{t}^{x} & = & [\begin{array}{ccc}
r_{t}^{y} & \cdots & r_{t+N-1}^{y}\end{array}]^{T},\label{eq:y_acc_time}\\
\mathbf{s}_{t}^{z} & = & [\begin{array}{ccc}
r_{t}^{z} & \cdots & r_{t+N-1}^{z}\end{array}]^{T}.\label{eq:z_acc_time}
\end{eqnarray}
The sequence size $N$ should be carefully selected such as to ensure an adequate and efficient activity recognition. We assume that the system supports $M$ different activities. Specifically, let $\mathcal{A}=\left\{ a_{1},a_{2},\ldots,a_{M}\right\}$ be a finite activity space. Based the windowed excerpts $\mathbf{s}_{t}^{x}$, $\mathbf{s}_{t}^{x}$, and $\mathbf{s}_{t}^{z}$, the activity recognition method infers the occurrence of an activity $y_{t}\in\mathcal{A}$.


\subsection{Data Preprocessing}

A \emph{spectrogram} of an accelerometer signal is a three dimensional representation of changes in the acceleration energy content of a signal as a function of frequency and time. Historically, spectrograms of speech waveforms are widely used as distinguishable features in acoustic modeling, e.g., the mel-frequency cepstral~\cite{zheng2001comparison}. In this paper, we use the spectrogram representation as the input of deep activity recognition models as it introduces the following advantages:
\begin{enumerate}
\item \textbf{Classification} \textbf{accuracy}. The spectrogram representation provides interpretable features in capturing the intensity differences among nearest acceleration data points. This enables the classification of activities based on the variations of spectral density which reduce the classification complexity.
\item \textbf{Computational complexity}. After applying the spectrogram on $\mathbf{s}_{t}^{x}$, $\mathbf{s}_{t}^{x}$, and $\mathbf{s}_{t}^{z}$, the length of the spectral signal is $L=3(\frac{N}{2}+1)$ while the time domain signal length is $3N$. This significantly reduces the computational burdens of any classification method due to the lower data dimensionality.
\end{enumerate}
Henceforth, the spectrogram signal of the triaxial accelerometer is denoted as $\mathbf{x}_{t}\in\mathbb{R}^{L}$, where $L=3(\frac{N}{2}+1)$ is the concatenated spectrogram signals from the triaxial input data.

\section{Deep Learning for Activity Recognition: System and Model}\label{sec:deep_learning_activity_recog}

Our deep model learns not only the classifier's weights used to recognize different activities, but also the informative features for recognizing these activities from raw data. This provides a competitive advantage over traditional systems that are hand-engineered. The model fitting and training consist of two main stages: (i)~An unsupervised, generative, and pre-training step, and (ii)~a supervised, discriminative, and fine-tuning step. The pre-training step generates intrinsic features based on a layer-by-layer training approach using unlabeled acceleration samples only. Firstly, we use deep belief networks~\cite{hinton2006fast} to find the activity membership probabilities. Then, we show how to utilize the activity membership probabilities generated by deep models to model the temporal correlation of sequential activities. 

\begin{figure}
\begin{centering}
\includegraphics[width=0.7\columnwidth,trim=1cm 1.5cm 1cm 0cm]{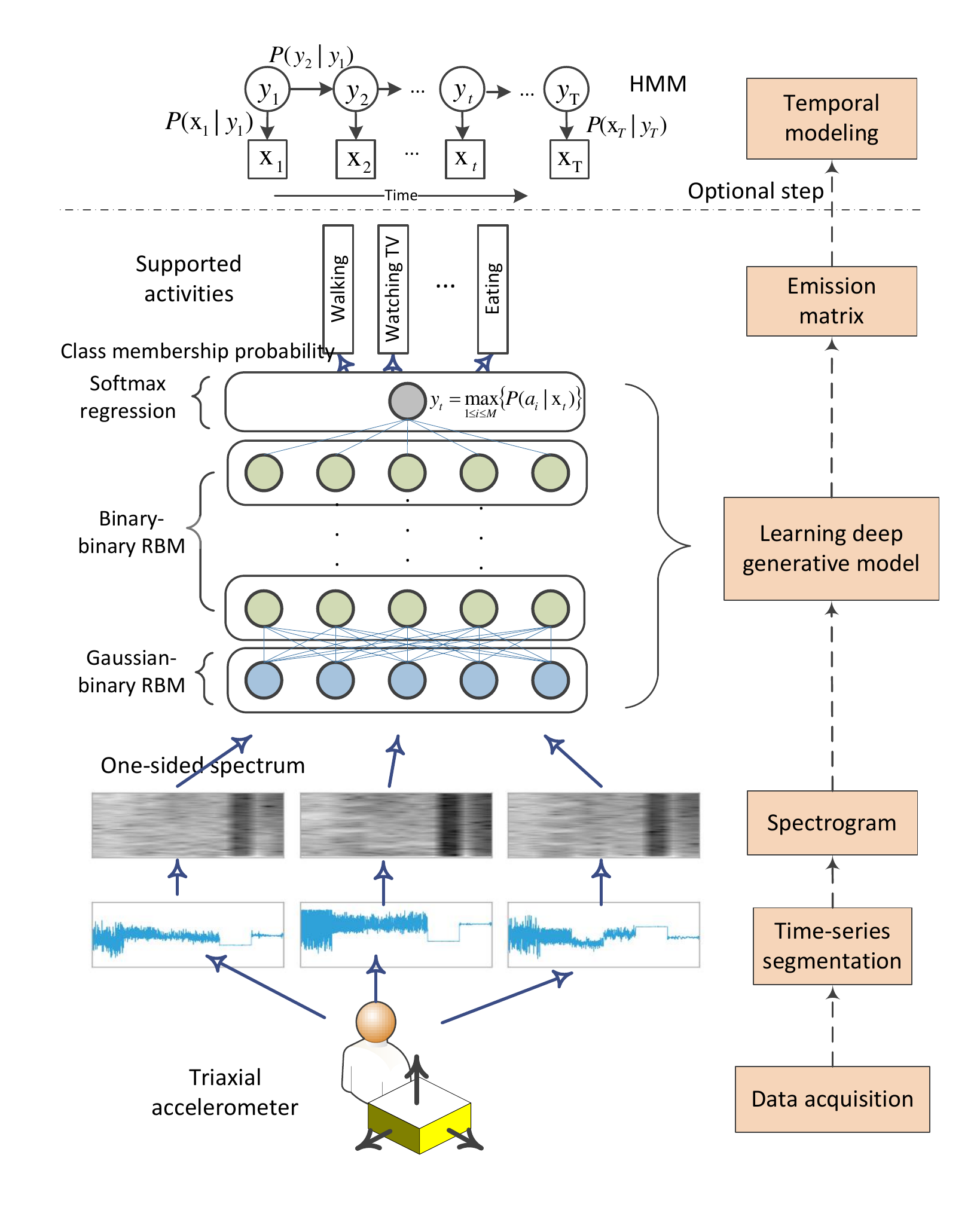}
\par\end{centering}

\caption{\textbf{Activity recognition using deep activity recognition model}. Our system automatically (1)~takes triaxial acceleration time series, (2)~extracts the spectrogram of windowed excerpts, (3)~computes intrinsic features using a deep generative model, and then (4)~recognizes the underlying human activities by finding the posterior probability distribution $\left\{ P\left(a_{i}|\mathbf{x}_{t}\right)\right\} _{i=1}^{M}$. This deep architecture outperforms existing methods for human activity recognition using accelerometers as shown by the experimental analysis on real world datasets. Furthermore, an optional step involves using the emission probabilities out of the deep model to train a hidden Markov model (HMM) for modeling temporal patterns in activities.\label{fig:system_chart}}
\end{figure}

Figure~\ref{fig:system_chart} shows the working flow of the proposed activity recognition system. We implement deep activity recognition models based on deep belief networks (BBNs). DBNs are generative models composed of multiple layers of hidden units. In~\cite{hinton2006fast}, the hidden units are formed from restricted Boltzmann machines (RBMs) which are trained in a layer-by-layer fashion. Notably, an alternative approach is based on using stacked auto-encoders~\cite{bengio2007greedy}. An RBM is a bipartite graph that is restricted in that no weight connections exist between hidden units. This restriction facilitates the model fitting as the hidden units become conditional independent for a given input vector. After the unsupervised pre-training, the learned weights are fine-tuned in an up-down manner using available data labels. A practical tutorial on the training of RBMs is presented in~\cite{hinton2012practical}.

\subsection{Deep Activity Recognition Models}\label{sub:deep_recognition}

\begin{figure}
\begin{centering}
\includegraphics[width=0.65\columnwidth,trim=1cm 1.5cm 1cm 0cm]{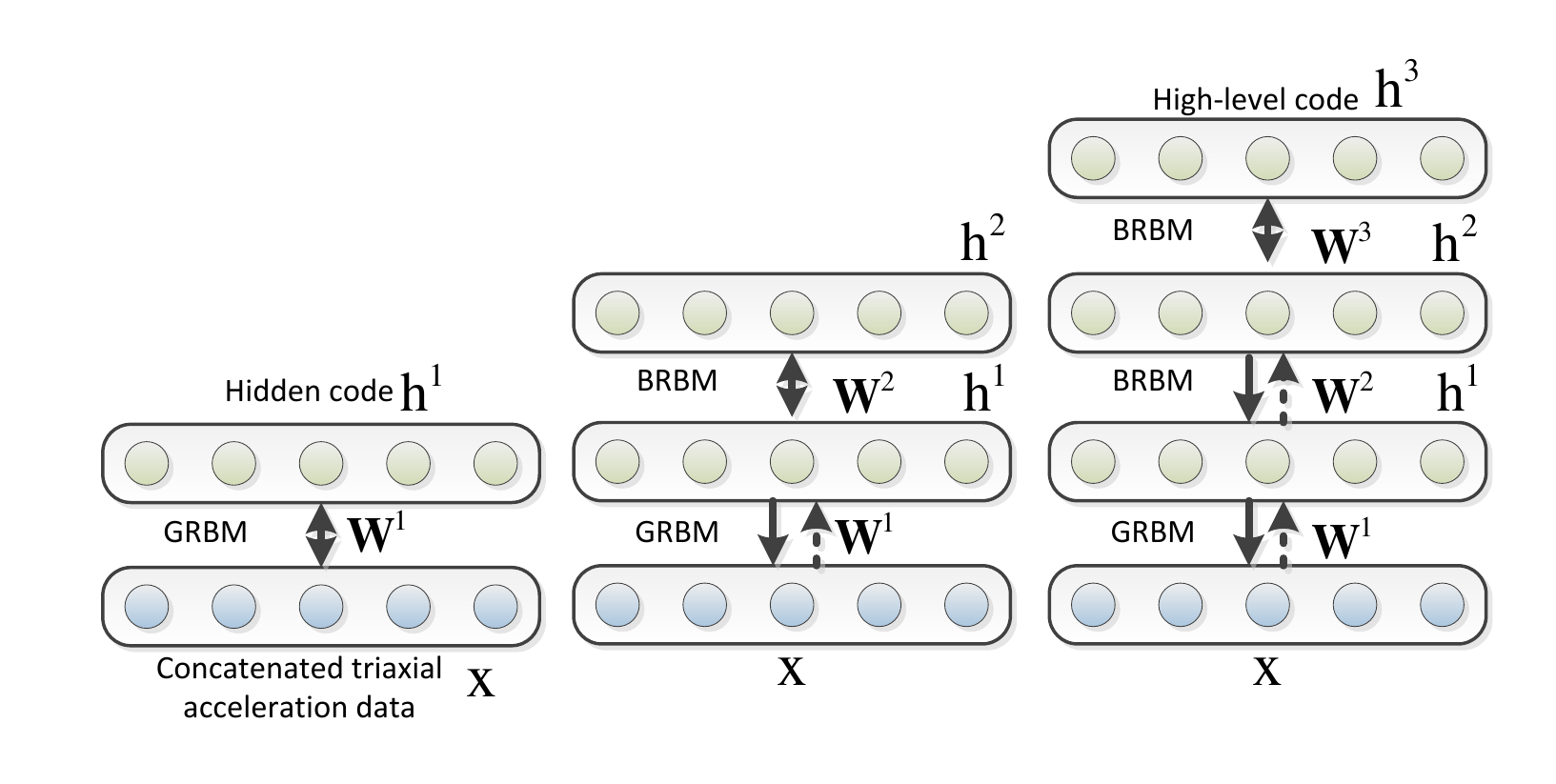}
\par\end{centering}

\caption{\textbf{The greedy layer-wise training of DBNs}. The first level is trained on triaxial acceleration data. Then, more RBMs are repeatedly stacked to form a deep activity recognition model until forming a high-level representation. \label{fig:greedy_layerwise_training}}
\end{figure}

DBNs~\cite{hinton2006fast} can be trained on greedy layer-wise training of RBMs as shown in Figure~\ref{fig:greedy_layerwise_training}. In our model, the acceleration spectrogram signals $\mathbf{x}$ are continuous and are fed to a deep activity recognition model. As a result, the first layer of the deep model is selected as a Gaussian-binary RBM (GRBM) which can model the energy content in the continuous accelerometer data. Afterward, the subsequent layers are binary-binary RBMs (BRBMs). RBMs are energy-based probabilistic models which are trained using stochastic gradient descent on the negative log-likelihood of the training data. For the GRBM layer, the energy of an observed vector $\mathbf{v}=\mathbf{x}$ and a hidden code $\mathbf{h}$ is denoted as follows:
\begin{equation}
E\left(\mathbf{v}=\mathbf{x},\mathbf{h}\right)=\frac{1}{2}\left(\mathbf{v}-\mathbf{b}\right)^{\top}\left(\mathbf{v}-\mathbf{b}\right)-\mathbf{c}^{\text{\ensuremath{\top}}}\mathbf{h}-\mathbf{v}^{\top}\mathbf{W}\mathbf{h}
\end{equation}
where $\mathbf{W}$ is the weight matrix connecting the input and hidden layers, $\mathbf{b}$ and $\mathbf{c}$ are the visible and hidden unit biases, respectively. For a BRBM, the energy function is defined as follows: 
\begin{equation}
E\left(\mathbf{v},\mathbf{h}\right)=-\mathbf{b}^{\text{\ensuremath{\top}}}\mathbf{v}-\mathbf{c}^{\text{\ensuremath{\top}}}\mathbf{h}-\mathbf{v}^{\text{\ensuremath{\top}}}\mathbf{W}\mathbf{h}.
\end{equation}
An RBM can be trained using the contrastive divergence approximation~\cite{hinton2002training} as follows:
\begin{equation}
\triangle W_{ij}=\alpha\left(\left\langle v_{i}h_{j}\right\rangle _{\text{data}}-\left\langle v_{i}h_{j}\right\rangle _{\text{1}}\right)
\end{equation}
where $\alpha$ is a learning rate. $\left\langle v_{i}h_{j}\right\rangle _{\text{data}}$ is the expectation of reconstruction over the data, and $\left\langle v_{i}h_{j}\right\rangle _{\text{1}}$is the expectation of reconstruction over the model using one step of the Gibbs sampler. Please refer to~\cite{hinton2006fast,hinton2012practical} for further details on the training of DBNs. For simplicity, we denote the weights and biases of a DBN model as $\mathbf{\theta}$ which can be used to find the posterior probabilities $P\left(a_{i}|\mathbf{x}_{t},\mathbf{\theta}\right)$ for each joint configuration $(a_{i},\mathbf{x}_{t})$.

To this end, the underlying activity $y_{t}$ can be predicted at time $t$ using the softmax regression as follows:
\begin{equation}
y_{t}=\arg\max_{1\leq i\leq M}\left\{ P\left(a_{i}|\mathbf{x}_{t},\mathbf{\theta}\right)\right\} .
\end{equation}
Alternatively, the temporal patterns in a sequence of activities can be further analyzed using HMMs. The following section establishes the probabilistic connection between the input data $\mathbf{x}_{t}$ and activity prediction $y_{t}$ over a sequence of observations $1\leq t\leq T$.

\subsection{Temporal Activity Recognition Models (DL-HMM)}\label{sub:temporal_recognition}

In some activity recognition applications, there is a temporal pattern in executed human activities, e.g., car checkpoint~\cite{zappi2007activity}. Hidden Markov models (HMMs)~\cite{rabiner1986introduction} are a type of graphical models that can simulate the temporal generation of a first-order Markov process. The temporal activity recognition problem includes finding the most probable sequence of (hidden) activities $y_{1},\ldots,y_{T}$ that produce an (observed) sequence of input $\mathbf{x}_{1},\ldots,\mathbf{x}_{T}$. An HMM model $\Phi$ is represented as a 3-tuple $\Phi=\left(\pi,\psi,\Upsilon\right)$ where $\pi=\left(P\left(y_{1}=a_{i}\right):i=1,\ldots,M\right)$ is the prior probabilities of all activities in the first hidden state, $\psi=\left(P\left(y_{t}=a_{i}|y_{t-1}=a_{j}\right):i,j=1,\ldots,M\right)$ is the transition probabilities, and $\Upsilon=\left(P\left(\mathbf{x}_{t}|y_{t}=a_{i}\right):i=1,\ldots,M\text{ and }t=1,\ldots,T\right)$ is the emission matrix for observables $\mathbf{x}_{t}$ from hidden symbols $a_{i}$. Given a sequence of observations, the emission probabilities is found using a deep model. In particular, the joint probabilities $P\left(y_{t},\mathbf{x}_{t}\right)$ of each joint configuration $(y_{t},\mathbf{x}_{t})$ in an HMM is found as follows:
\begin{eqnarray}
P\left(y_{t},\mathbf{x}_{t}\right) &= & P\left(y_{1}\right)P\left(\mathbf{x}_{1}|y_{1}\right)\prod_{i=2}^{T}P\left(y_{i}|y_{i-1}\right)P\left(\mathbf{x}_{i}|y_{i}\right),\\
 & = & P\left(y_{t-1},\mathbf{x}_{t-1}\right)P\left(y_{t}|y_{t-1}\right)P\left(\mathbf{x}_{t}|y_{t}\right),\label{eq:recursive_hmm}
\end{eqnarray}
Herein, (\ref{eq:recursive_hmm}) shows that an HMM infers the posterior distribution $P\left(y_{t}|\mathbf{x}_{t}\right)$ as a recursive process. This decoding problem is solved for the most probable path of sequential activities.

\subsection{Computational Complexity}

Our algorithm consists of three working phases: (a)~data gathering, (b)~offline learning, and (c)~online activity recognition and inference. The computational burden of the offline learning is relatively heavy to be run on a mobile device as it based on stochastic gradient descent optimization. Therefore, it is recommended to run the offline training of a deep activity recognition model on a capable server. Nonetheless, after the offline training is completed, the model parameter $\mathbf{\theta}$ is only disseminated to the wearable device where the online activity recognition is lightweight with a linear time complexity ($\mathcal{O}\left(L\times D\right)$), where $D$ is the number of layers in the deep activity recognition model. Here, the time complexity of the online activity recognition system represents the time needed to recognize the activity as a function of the accelerometer input length. The time complexity of finding the short-time Fourier transform (STFT) is $\mathcal{O}\left(L\log\left(L\right)\right)$. Finally, the time complexity of the HMM decoding problem is $\mathcal{O}\left(M^{2}\times T\right)$.

\section{Baselines and Result Summary}\label{sec:baselines_results}


\subsection{Datasets}

For empirical comparison with existing approaches, we use three public datasets that represent different application domains to verify the efficiency of our proposed solution. These three testbeds are described as follows:
\begin{itemize}
\item \textbf{WISDM Actitracker dataset}~\cite{kwapisz2011activity}: This dataset contains $1,098,213$ samples of one triaxial accelerometer that is programmed to sample at a rate of $20$ Hz. The data samples belong to $29$ users and $6$ distinctive human activities of walking, jogging, sitting, standing, and climbing stairs. The acceleration samples are collected using mobile phones with Android operating system.
\item \textbf{Daphnet freezing of gait dataset}~\cite{bachlin2010wearable}: We used this dataset to demonstrate the healthcare applications of deep activity recognition models. The data samples are collected from patients with the Parkinson's disease. Three triaxial accelerometers are fixed at patient's ankle, upper leg, and trunk with a sampling frequency of $64$ Hz. The objective is to detect freezing events of patients. The dataset contains $1,140,835$ experimentation samples from $10$ users. The samples are labeled with either ``freezing'' or ``no freezing'' classes.
\item \textbf{Skoda checkpoint dataset}~\cite{zappi2007activity}: The $10$ distinctive activities of this dataset belong to a car maintenance scenario in typical quality control checkpoints. The sampling rate is $98$~Hz. Even though the dataset contains $20$ nodes of triaxial accelerometers, it would be inconvenient and costly to fix $20$ nodes to employee hands which can hinder the maintenance work. Therefore, we use one accelerometer node (ID \# 16) for the experimental validation of deep models.
\end{itemize}

\subsection{Performance Measures}

For binary classification (experimentation on the Daphnet dataset), we use three performance metrics: $\text{Sensitivity (TPR)}=\frac{\textrm{TP}}{\textrm{TP+FN}}$, $\text{specificity (TNR)}=\frac{\textrm{TN}}{\textrm{TN+FP}}$, and $\text{accuracy (ACC)}=\frac{\textrm{TP+TN}}{\textrm{TP+TN+FP+FN}}$ where TP, TN, FP, and FN mean true positive, true negative, false positive, and false negative, respectively. For multiclass classification of non-overlapping activities, which are based on the experimentation of the WISDM Actitracker and Skoda checkpoint datasets, the average recognition accuracy (ACC) is found as $\textrm{ACC}=\frac{1}{M}\sum_{i=1}^{M}\frac{TP_{i}+TN_{i}}{TP_{i}+TN_{i}+FP_{i}+FN_{i}}$, where $M$ is the number of supported activities.

\subsection{Baselines}

Table~\ref{tab:result_summay} summarizes the main performance results of our proposed method and some previous solutions on using the three datasets. Deep activity recondition models introduce significant accuracy improvement over conventional methods. For example, it improves accuracy by $6.53\%$ over MLPs and $3.93\%$ over ensemble learning on the WISDM Actitracker dataset. Similarly, significant improvements are also reported for the Daphnet freezing of gait and\textbf{ }Skoda  checkpoint datasets. This summarized result shows that the deep models are both (a)~effective in improving recognition accuracy over state-of-the-art methods, and (b)~practical for avoiding the hand-engineering of features. 

\begin{table*}
\caption{Comparison of our proposed solution against existing methods in terms
of recognition accuracy. C4.5 is a decision tree generation method.\label{tab:result_summay}}

\centering{}%
\begin{tabular}{|c|c|>{\centering}p{3.5cm}|>{\centering}m{2cm}|>{\centering}p{2.5cm}|}
\hline 
\textbf{\noun{Dataset}} & \textbf{\noun{Reference}} & \textbf{\noun{Solution}} & \textbf{\noun{Window size}} & \textbf{\noun{Accuracy (\%)}}\tabularnewline
\hline 
\hline 
\multirow{5}{*}{WISDM} & \cite{kwapisz2011activity} & C4.5 & \multirow{5}{2cm}{\centering{}10 sec} & 85.1\tabularnewline
\cline{2-3} \cline{5-5} 
 & \cite{kwapisz2011activity} & Logistic regression &  & 78.1\tabularnewline
\cline{2-3} \cline{5-5} 
 & \cite{kwapisz2011activity} & MLPs &  & 91.7\tabularnewline
\cline{2-3} \cline{5-5} 
 & \cite{catal2015use} & Ensemble learning &  & 94.3\tabularnewline
\cline{2-3} \cline{5-5} 
 & Our solution & Deep learning models &  & \textbf{98.23}\tabularnewline
\hline 
\multirow{3}{*}{Daphnet} & \cite{bachlin2010wearable} & Energy threshold on power spectral density (0.5sec) & 4 sec & TPR: 73.1 and

TNR: 81.6\tabularnewline
\cline{2-5} 
 & \cite{hammerla2013preserving} & C4.5 and k-NNs with feature extraction methods & - & TPR and TNR $\sim$ 82\tabularnewline
\cline{2-5} 
 & Our solution & Deep learning models & 4 sec & \textbf{TPR and TNR $\sim$ 91.5}\tabularnewline
\hline 
\multirow{2}{*}{Skoda} & \cite{zappi2007activity} & HMMs & - & Node 16 (86),

nodes 20, 22 and 25 (84)\tabularnewline
\cline{2-5} 
 & Our solution & Deep learning models & 4 sec & Node 16 (\textbf{89.38})\tabularnewline
\hline 
\end{tabular}

\end{table*}

\section{Experiments on Real Datasets}\label{sec:experiments}


\subsection{Spectrogram Analysis}\label{sub:data_analysis}

\begin{figure}
\begin{centering}
\includegraphics[width=1\columnwidth,trim=1cm 1.5cm 1cm 0cm]{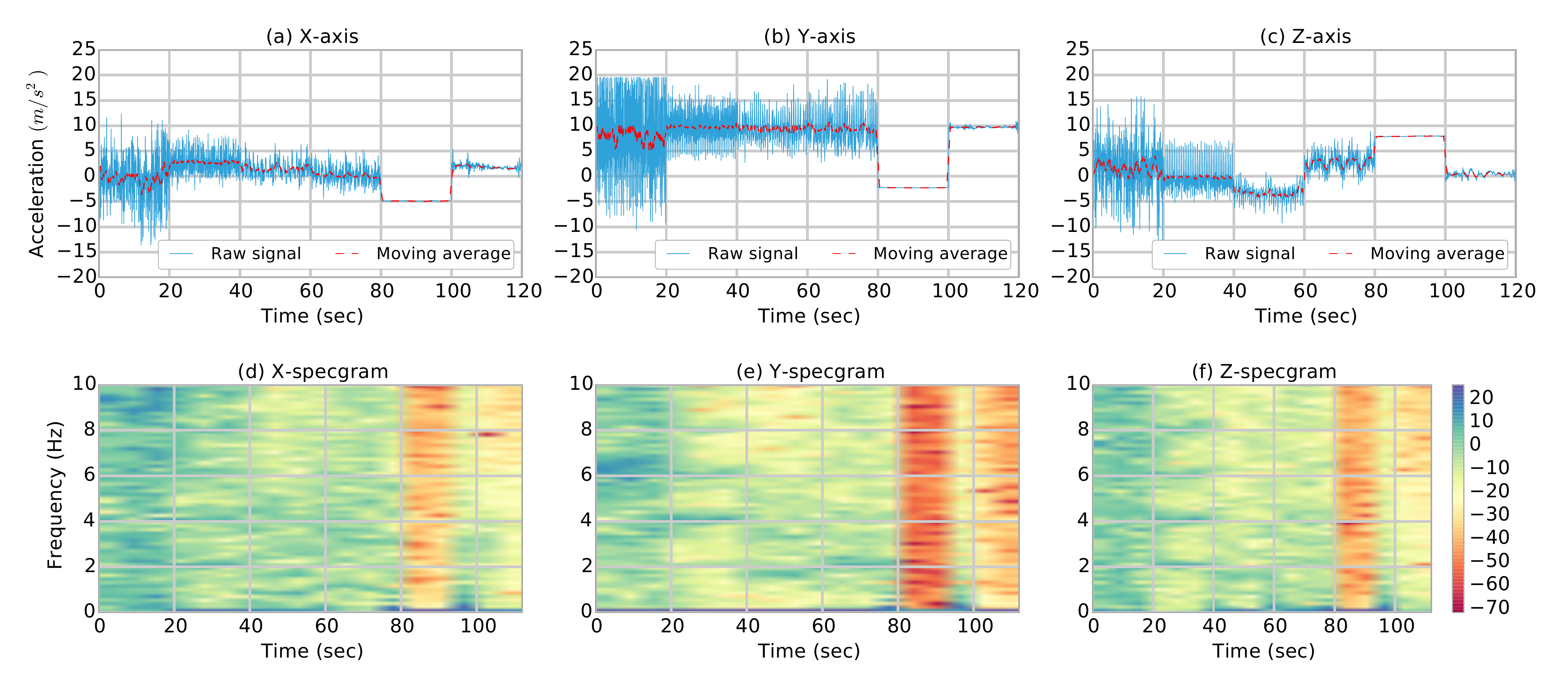}
\par\end{centering}

\caption{\textbf{Frequency spectrum as a parametric representation}. Data samples of a triaxial accelerometer and their corresponding spectrogram representation (WISDM Actitracker dataset). These samples belong to five everyday human activities: jogging $t\in[0,20)$, walking $t\in[20,40)$, upstairs $t\in[40,60)$, downstairs $t\in[60,80)$, sitting $t\in[80,100)$, and standing $t\in[100,120)$. The acceleration signal is usual subtle and only cover a small range of the frequency domain.\label{fig:wisdm_data}}
\end{figure}

Figure~\ref{fig:wisdm_data} shows triaxial time series and spectrogram signals of 6 activities of the WISDM Actitracker dataset. Clearly, the high frequency signals (a.k.a. AC components) belong to activities with active body motion, e.g., jogging and walking. On the other hand, the low frequency signals (a.k.a. DC components) are collected during semi-static body motions, e.g., sitting and standing. Thereby, these low frequency activities are only distinguishable by the accelerometer measurement of the gravitational acceleration.

\subsection{Performance Analysis}\label{sub:performance_analysis}


In our experiments, the data is firstly centered to the mean and scaled to a unit variance. The deep activity recognition models are trained using stochastic gradient decent with mini-batch size of $75$. For the first GBRM layer, the pre-training learning rate is set to $0.001$ with pre-training epochs of $150$. For next BRBM layers, the number of pre-training epochs is fixed to $75$ with pre-training learning rate of $0.01$. The fine-tuning learning rate is $0.1$ and the number of fine-tuning epochs is $1000$. For interested technical readers, Hinton~\cite{hinton2012practical} provides a tutorial on training RBMs with many practical advices on parameter setting and tuning.

\subsubsection{Deep Model Structure}

\begin{figure}
\begin{centering}
\includegraphics[width=0.9\columnwidth,trim=1cm 1.5cm 1cm 0cm]{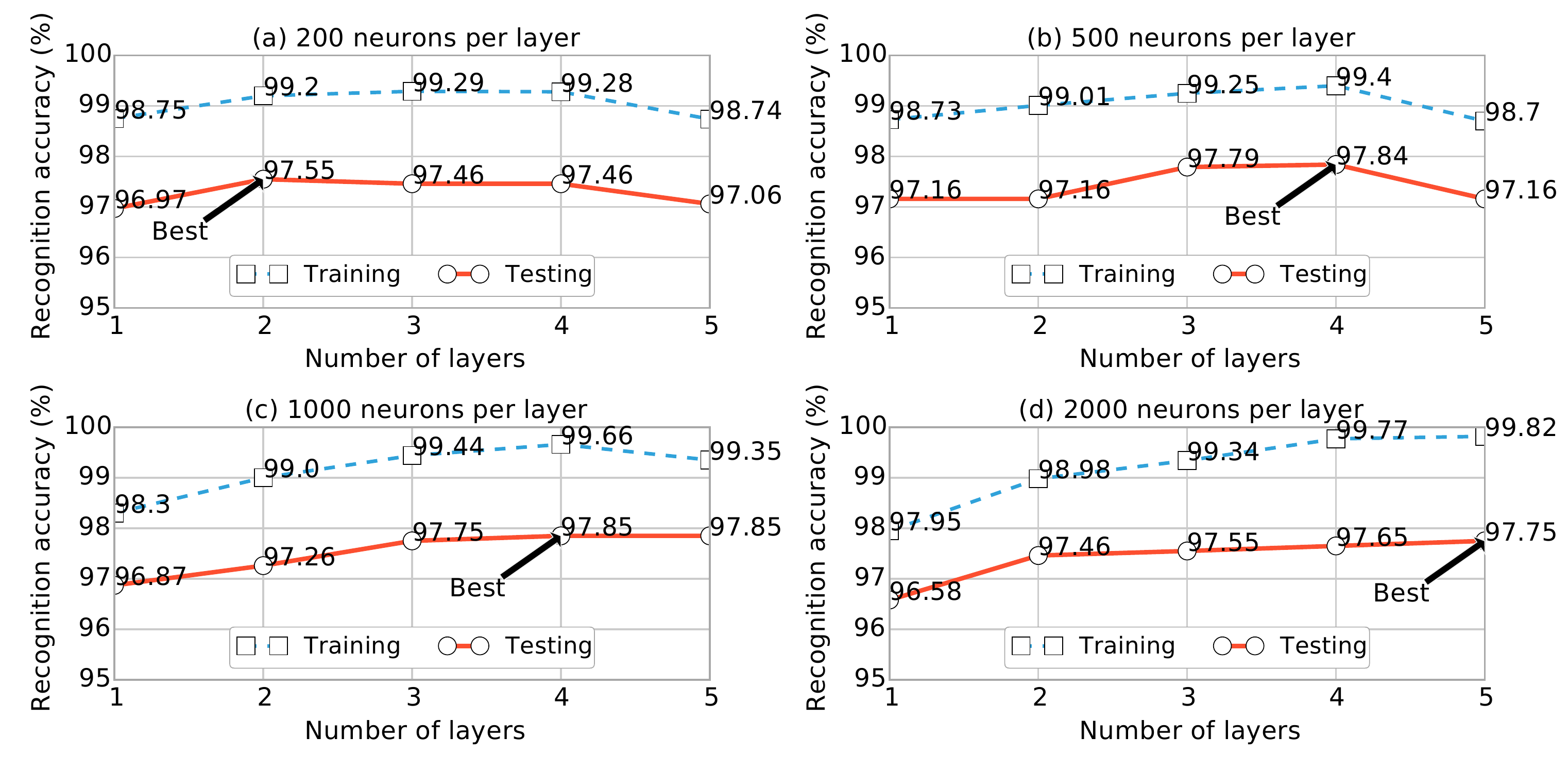}
\par\end{centering}

\caption{\textbf{Optimizing deep activity recognition models}. Activity recognition using the WISDM Actitracker dataset under different DBN setup. At each figure, the rates of activity recognition accuracy are shown for both training and testing data samples. The input length is 303 which corresponds to 10-second frames.\label{fig:wisdm_setup}}
\end{figure}

Figure~\ref{fig:wisdm_setup} shows the recognition accuracy on different DBN structures (joint configurations of number of layers and number of neurons per layer). Two important results are summarized as follows:
\begin{enumerate}
\item \textbf{Deep models outperforms shallow ones}. Clearly, the general trend in the recognition accuracy is that using more layers will enhance the recognition accuracy. For example, using $4$ layers of $500$ neurons at each layer is better than $2$ layers of $1000$ neurons at each layer, which is better than 1 layer of $2000$ neurons.
\item \textbf{Overcomplete representations are advantageous}. An overcompete representation is achieved when the number of neurons at each layer is larger than the input length. An overcompete representation is essential for learning deep models with many hidden layers, e.g., deep model of $2000$ neurons per layer. On the other hand, it is noted that a deep model will be hard to optimized when using undercomplete representations, e.g., $5$ layers of $200$ neurons at each layer. This harder optimization issue is distinguishable from the overfitting problem as the training data accuracy is also degrading by adding more layers (i.e., an overfitted model is diagnosed when the recognition accuracy on training data is enhancing by adding more layer while getting poorer accuracy on testing data). Therefore, we recommend 4x overcomplete deep activity recognition models (i.e., the number of neurons at each layer is four times the input size).
\end{enumerate}

\subsubsection{Pre-training Effects}

\begin{table}
\begin{centering}
\caption{Comparison of accuracy improvements due to the pre-training stage. Each layer consists of 1000 neurons. \label{tab:pretraining_effect}}
\begin{tabular}{|>{\centering}m{3cm}|c|>{\centering}p{2cm}|}
\hline 
\textbf{\noun{Expriement}} & \textbf{\noun{\# of layers}} & \textbf{\noun{Accuracy (\%)}}\tabularnewline
\hline 
\hline 
\multirow{3}{3cm}{Generative \& discriminative training} & 1 & 96.87\tabularnewline
\cline{2-3} 
 & 3 & 97.75\tabularnewline
\cline{2-3} 
 & 5 & 97.85\tabularnewline
\hline 
\multirow{3}{3cm}{Discriminative training only} & 1 & 96.87\tabularnewline
\cline{2-3} 
 & 3 & 96.46\tabularnewline
\cline{2-3} 
 & 5 & 96.51\tabularnewline
\hline 
\end{tabular}
\par\end{centering}

\end{table}

Table~\ref{tab:pretraining_effect} shows the recognition accuracy with and without the pre-training phase. These results confirm the importance of the generative pre-training phase of deep activity recognition models. Specifically, a generative pre-training of a deep model guides the discriminative training to better generalization solutions~\cite{erhan2010does}. Clearly, the generative pre-training is almost ineffective for 1-layer networks. However, using the generative pre-training becomes more essential for the recognition accuracy of deeper activity recognition models, e.g., 5 layers.

\subsection{Temporal Modeling}\label{sub:temporal_experiment}

\begin{figure}
\begin{centering}
\includegraphics[width=0.75\columnwidth,trim=1cm 1.5cm 1cm 0cm]{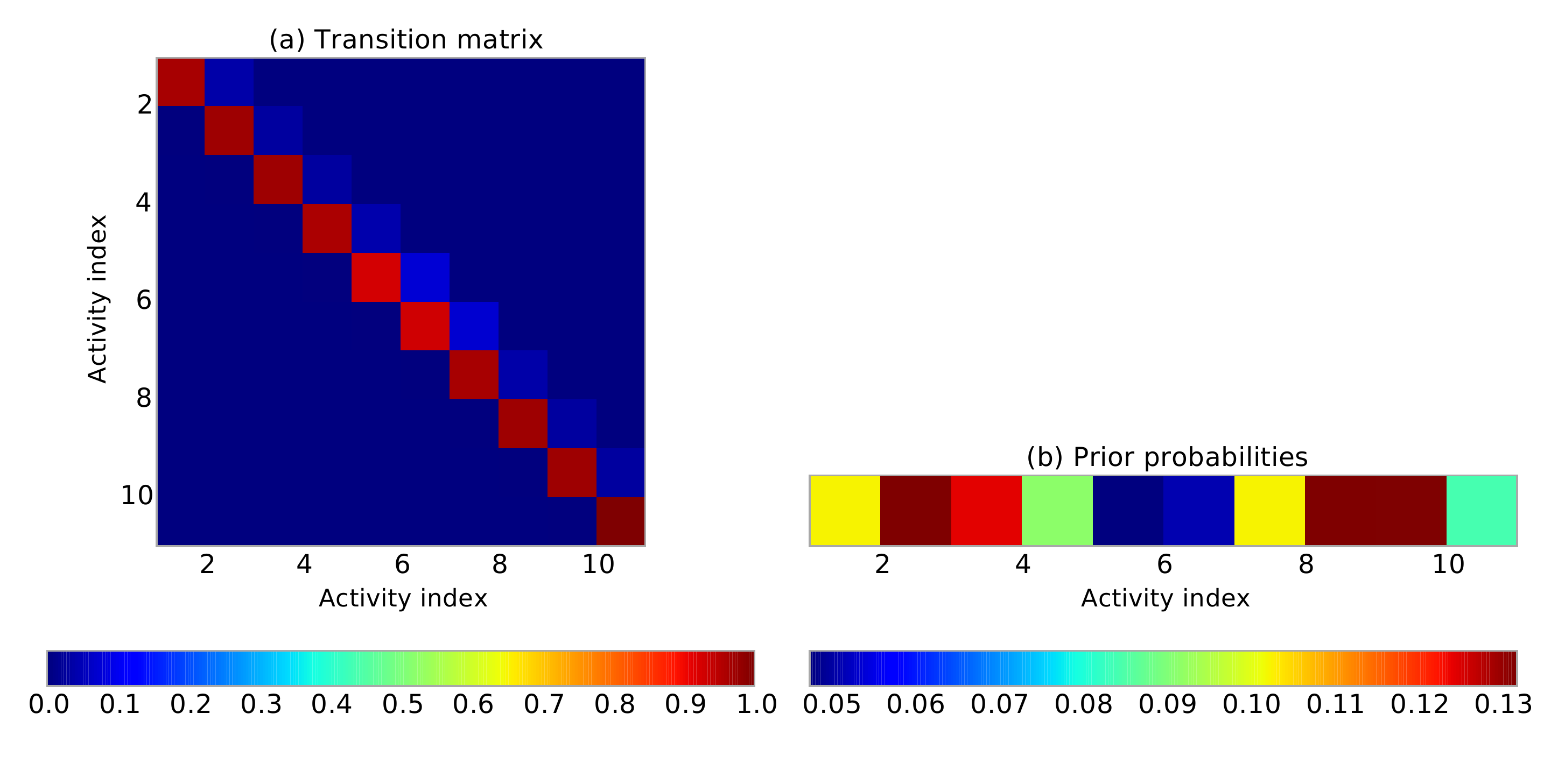}
\par\end{centering}

\caption{\textbf{Transition and prior probabilities of sequential activities}. (a)~The transition matrix $\psi\in\mathbb{R}^{N\times N}$ that represents the probabilities of moving among activities $\mathcal{A}=\left\{ a_{1},a_{2},\ldots,a_{10}\right\}$. (b)~The prior belief $\pi\in\mathbb{R}^{N}$ that stores the initial probabilities of different activities. These parameters are extracted from the Skoda  checkpoint dataset (node ID 16).\label{fig:skoda_temporal_modeling}}
\end{figure}

We used a deep activity recognition model with 3 layers of 1000 neurons each. The recognition accuracy is $89.38\%$ for the 10 activities on the Skoda checkpoint dataset (node ID 16), improving $3.38\%$ over the HMM method presented by \cite{zappi2007activity}. Furthermore, the results can be significantly enhanced by exploring the temporal correlation in the dataset. Our hybrid DL-HMM achieves near perfect recognition accuracy of $99.13\%$. In particular, Figure~\ref{fig:skoda_temporal_modeling} shows the parameters of a HMM model that is used to model the temporal sequences of the Skoda checkpoint dataset. Here, the checkpoint task must follow a specific activity sequence.

\section{Conclusions and Future Work}\label{sec:conclusions}

We investigated the problem of activity recognition using triaxial accelerometers. The proposed approach is superior to traditional methods of using shallow networks with handcrafted features by using deep activity recognition models. The deep activity recognition models produce significant improvement to the recognition accuracy by extracting hierarchical features from triaxial acceleration data. Moreover, the recognition probabilities of deep activity recognition models are utilized as an emission matrix of a hidden Markov model to temporally model a sequence of human activities.


\bibliography{references}

\begin{thebibliography}{}

\bibitem[\protect\citeauthoryear{Altun and Barshan}{2010}]{altun2010human}
Altun, K., and Barshan, B.
\newblock 2010.
\newblock Human activity recognition using inertial/magnetic sensor units.
\newblock In {\em Human Behavior Understanding}. Springer.
\newblock  38--51.

\bibitem[\protect\citeauthoryear{B{\"a}chlin \bgroup et al\mbox.\egroup
  }{2010}]{bachlin2010wearable}
B{\"a}chlin, M.; Plotnik, M.; Roggen, D.; Maidan, I.; Hausdorff, J.~M.; Giladi,
  N.; and Tr{\"o}ster, G.
\newblock 2010.
\newblock Wearable assistant for {P}arkinson's disease patients with the
  freezing of gait symptom.
\newblock {\em IEEE Transactions on Information Technology in Biomedicine}
  14(2):436--446.

\bibitem[\protect\citeauthoryear{Bengio \bgroup et al\mbox.\egroup
  }{2007}]{bengio2007greedy}
Bengio, Y.; Lamblin, P.; Popovici, D.; Larochelle, H.; et~al.
\newblock 2007.
\newblock Greedy layer-wise training of deep networks.
\newblock {\em Advances in neural information processing systems} 19:153.

\bibitem[\protect\citeauthoryear{Berchtold \bgroup et al\mbox.\egroup
  }{2010}]{berchtold2010actiserv}
Berchtold, M.; Budde, M.; Gordon, D.; Schmidtke, H.~R.; and Beigl, M.
\newblock 2010.
\newblock Acti{S}erv: {A}ctivity recognition service for mobile phones.
\newblock In {\em Proceedings of the International Symposium on Wearable
  Computers},  1--8.
\newblock IEEE.

\bibitem[\protect\citeauthoryear{Catal \bgroup et al\mbox.\egroup
  }{2015}]{catal2015use}
Catal, C.; Tufekci, S.; Pirmit, E.; and Kocabag, G.
\newblock 2015.
\newblock On the use of ensemble of classifiers for accelerometer-based
  activity recognition.
\newblock {\em Applied Soft Computing}.

\bibitem[\protect\citeauthoryear{Chen \bgroup et al\mbox.\egroup
  }{2012}]{chen2012sensor}
Chen, L.; Hoey, J.; Nugent, C.~D.; Cook, D.~J.; and Yu, Z.
\newblock 2012.
\newblock Sensor-based activity recognition.
\newblock {\em IEEE Transactions on Systems, Man, and Cybernetics, Part C:
  Applications and Reviews} 42(6):790--808.

\bibitem[\protect\citeauthoryear{Dahl \bgroup et al\mbox.\egroup
  }{2012}]{dahl2012context}
Dahl, G.~E.; Yu, D.; Deng, L.; and Acero, A.
\newblock 2012.
\newblock Context-dependent pre-trained deep neural networks for
  large-vocabulary speech recognition.
\newblock {\em IEEE Transactions on Audio, Speech, and Language Processing}
  20(1):30--42.

\bibitem[\protect\citeauthoryear{Erhan \bgroup et al\mbox.\egroup
  }{2010}]{erhan2010does}
Erhan, D.; Bengio, Y.; Courville, A.; Manzagol, P.-A.; Vincent, P.; and Bengio,
  S.
\newblock 2010.
\newblock Why does unsupervised pre-training help deep learning?
\newblock {\em The Journal of Machine Learning Research} 11:625--660.

\bibitem[\protect\citeauthoryear{Fender \bgroup et al\mbox.\egroup
  }{2008}]{fender2008two}
Fender, A.; MacPherson, W.~N.; Maier, R.; Barton, J.~S.; George, D.~S.; Howden,
  R.~I.; Smith, G.~W.; Jones, B.; McCulloch, S.; Chen, X.; et~al.
\newblock 2008.
\newblock Two-axis temperature-insensitive accelerometer based on multicore
  fiber {B}ragg gratings.
\newblock {\em IEEE sensors journal} 7(8):1292--1298.

\bibitem[\protect\citeauthoryear{Hammerla \bgroup et al\mbox.\egroup
  }{2013}]{hammerla2013preserving}
Hammerla, N.~Y.; Kirkham, R.; Andras, P.; and Ploetz, T.
\newblock 2013.
\newblock On preserving statistical characteristics of accelerometry data using
  their empirical cumulative distribution.
\newblock In {\em Proceedings of the International Symposium on Wearable
  Computers},  65--68.
\newblock ACM.

\bibitem[\protect\citeauthoryear{Hinton, Osindero, and
  Teh}{2006}]{hinton2006fast}
Hinton, G.~E.; Osindero, S.; and Teh, Y.-W.
\newblock 2006.
\newblock A fast learning algorithm for deep belief nets.
\newblock {\em Neural computation} 18(7):1527--1554.

\bibitem[\protect\citeauthoryear{Hinton}{2002}]{hinton2002training}
Hinton, G.~E.
\newblock 2002.
\newblock Training products of experts by minimizing contrastive divergence.
\newblock {\em Neural computation} 14(8):1771--1800.

\bibitem[\protect\citeauthoryear{Hinton}{2012}]{hinton2012practical}
Hinton, G.~E.
\newblock 2012.
\newblock A practical guide to training restricted {B}oltzmann machines.
\newblock In {\em Neural Networks: Tricks of the Trade}. Springer.
\newblock  599--619.

\bibitem[\protect\citeauthoryear{Khan \bgroup et al\mbox.\egroup
  }{2010}]{khan2010triaxial}
Khan, A.~M.; Lee, Y.-K.; Lee, S.~Y.; and Kim, T.-S.
\newblock 2010.
\newblock A triaxial accelerometer-based physical-activity recognition via
  augmented-signal features and a hierarchical recognizer.
\newblock {\em IEEE Transactions on Information Technology in Biomedicine}
  14(5):1166--1172.

\bibitem[\protect\citeauthoryear{Kwapisz, Weiss, and
  Moore}{2011}]{kwapisz2011activity}
Kwapisz, J.~R.; Weiss, G.~M.; and Moore, S.~A.
\newblock 2011.
\newblock Activity recognition using cell phone accelerometers.
\newblock {\em ACM SigKDD Explorations Newsletter} 12(2):74--82.

\bibitem[\protect\citeauthoryear{Lara and Labrador}{2013}]{lara2013survey}
Lara, O.~D., and Labrador, M.~A.
\newblock 2013.
\newblock A survey on human activity recognition using wearable sensors.
\newblock {\em IEEE Communications Surveys \& Tutorials} 15(3):1192--1209.

\bibitem[\protect\citeauthoryear{Mohamed, Dahl, and
  Hinton}{2012}]{mohamed2012acoustic}
Mohamed, A.-R.; Dahl, G.~E.; and Hinton, G.
\newblock 2012.
\newblock Acoustic modeling using deep belief networks.
\newblock {\em IEEE Transactions on Audio, Speech, and Language Processing}
  20(1):14--22.

\bibitem[\protect\citeauthoryear{Parkka \bgroup et al\mbox.\egroup
  }{2006}]{parkka2006activity}
Parkka, J.; Ermes, M.; Korpipaa, P.; Mantyjarvi, J.; Peltola, J.; and Korhonen,
  I.
\newblock 2006.
\newblock Activity classification using realistic data from wearable sensors.
\newblock {\em IEEE Transactions on Information Technology in Biomedicine}
  10(1):119--128.

\bibitem[\protect\citeauthoryear{Pl{\"o}tz, Hammerla, and
  Olivier}{2011}]{plotz2011feature}
Pl{\"o}tz, T.; Hammerla, N.~Y.; and Olivier, P.
\newblock 2011.
\newblock Feature learning for activity recognition in ubiquitous computing.
\newblock In {\em IJCAI Proceedings-International Joint Conference on
  Artificial Intelligence}, volume~22,  1729.

\bibitem[\protect\citeauthoryear{Rabiner and
  Juang}{1986}]{rabiner1986introduction}
Rabiner, L.~R., and Juang, B.-H.
\newblock 1986.
\newblock An introduction to hidden markov models.
\newblock {\em IEEE ASSP Magazine} 3(1):4--16.

\bibitem[\protect\citeauthoryear{Rumelhart, Hinton, and
  Williams}{1986}]{rumelhart1988learning}
Rumelhart, D.~E.; Hinton, G.~E.; and Williams, R.~J.
\newblock 1986.
\newblock Learning representations by back-propagating errors.
\newblock {\em Nature} 323(6088):533--536.

\bibitem[\protect\citeauthoryear{Salakhutdinov}{2015}]{salakhutdinov2015learning}
Salakhutdinov, R.
\newblock 2015.
\newblock Learning deep generative models.
\newblock {\em Annual Review of Statistics and Its Application} 2(1):361--385.

\bibitem[\protect\citeauthoryear{Xu \bgroup et al\mbox.\egroup
  }{2012}]{xu2012robust}
Xu, W.; Zhang, M.; Sawchuk, A.~A.; and Sarrafzadeh, M.
\newblock 2012.
\newblock Robust human activity and sensor location corecognition via sparse
  signal representation.
\newblock {\em IEEE Transactions on Biomedical Engineering} 59(11):3169--3176.

\bibitem[\protect\citeauthoryear{Zappi \bgroup et al\mbox.\egroup
  }{2008}]{zappi2007activity}
Zappi, P.; Lombriser, C.; Stiefmeier, T.; Farella, E.; Roggen, D.; Benini, L.;
  and Tr{\"o}ster, G.
\newblock 2008.
\newblock Activity recognition from on-body sensors: {A}ccuracy-power trade-off
  by dynamic sensor selection.
\newblock In {\em Wireless sensor networks}. Springer.
\newblock  17--33.

\bibitem[\protect\citeauthoryear{Zheng, Zhang, and
  Song}{2001}]{zheng2001comparison}
Zheng, F.; Zhang, G.; and Song, Z.
\newblock 2001.
\newblock Comparison of different implementations of {MFCC}.
\newblock {\em Journal of Computer Science and Technology} 16(6):582--589.

\end{thebibliography}
\bibliographystyle{aaai}

\end{document}